# TOWARDS MORE ACCURATE CLUSTERING METHOD BY USING DYNAMIC TIME WARPING

Khadoudja Ghanem

MISC Laboratory, Constantine 2 University , Nouvelle ville, Constantine, Algeria ;
gkhadoudja@yahoo.fr

*ABSTRACT*

*An intrinsic problem of classifiers based on machine learning (ML) methods is that their learning time grows as the size and complexity of the training dataset increases. For this reason, it is important to have efficient computational methods and algorithms that can be applied on large datasets, such that it is still possible to complete the machine learning tasks in reasonable time. In this context, we present in this paper a more accurate simple process to speed up ML methods. An unsupervised clustering algorithm is combined with Expectation, Maximization (EM) algorithm to develop an efficient Hidden Markov Model (HMM) training. The idea of the proposed process consists of two steps. In the first step, training instances with similar inputs are clustered and a weight factor which represents the frequency of these instances is assigned to each representative cluster. Dynamic Time Warping technique is used as a dissimilarity function to cluster similar examples. In the second step, all formulas in the classical HMM training algorithm (EM) associated with the number of training instances are modified to include the weight factor in appropriate terms. This process significantly accelerates HMM training while maintaining the same initial, transition and emission probabilities matrixes as those obtained with the classical HMM training algorithm. Accordingly, the classification accuracy is preserved. Depending on the size of the training set, speedups of up to 2200 times is possible when the size is about 100.000 instances. The proposed approach is not limited to training HMMs, but it can be employed for a large variety of MLs methods.*

## 1. INTRODUCTION

The main problem of classifiers based on machine learning (ML) methods is that their learning time grows as the size and complexity of the training dataset increases because training with only a few data will lead to an unreliable performance. Indeed, due to advances in technology, the size and dimensionality of data sets used in machine learning tasks have grown very large and continue to grow by the day. For this reason, it is important to have efficient computational methods and algorithms that can be applied on very large data sets, such that it is still possible to complete the machine learning tasks in reasonable time.

The speeding up of the estimation process of the built model parameters is a common problem of ML methods, although it raises several challenges. Many methods have been employed to overcome this problem. These methods can be classified into two categories: the first category consists of methods which compress data, either the number of instances [2, 3, 4, 5], or the set of features (attributes) which characterize instances [16,20,21]. The second category consists of methods which reduce running time in the execution and compilation level [16, 22, 23].

DOI : 10.5121/ijdkp.2013.3207 107

This paper is an extended version of our previous work described in [1]. In that paper a simple process to speed up ML methods is presented. An unsupervised clustering algorithm is combined with Expectation, Maximization (EM) algorithm to develop an efficient Hidden Markov Model (HMM) training. The idea of the proposed process consists of two steps. The first one involves a preprocessing step to reduce the number of training instances with any information loss. In this step, training instances with similar inputs are clustered and a weight factor which represents the frequency of these instances is assigned to each representative cluster (index). The Euclidean distance is used as a metric to compare two observation sequences. Increasing evidence with the use of this metric is its poor accuracy for classification and clustering of time dependent sequences. The Euclidean distance metric is widely known to be very sensitive to distortion in time axis [6][7]. In spite of that it is used in many fields because of its ease of implementation and its time and space efficiency.

In this paper, we introduce Dynamic Time Warping (DTW) as a dissimilarity function. The problem of distortion in the time axis can be addressed by DTW. This method allows non-linear alignments between two sequences with different length to accommodate those that are similar, but locally out of phase, as shown in Figure 1.

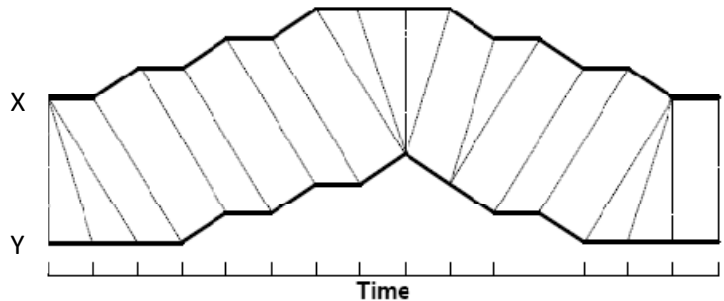

Figure 1, a warping between two temporal signals (Matching by stretching & scaling along time axis).
.

The Euclidean distance between two sequences can be seen as a special case of DTW, it is only defined in the special case where the two sequences have the same length. It is a way to solve a vast range of time dependant sequence problems, and it is widely used in various disciplines:- In bioinformatics, Aach and Church successfully applied DTW to cluster RNA expression data [8]. In chemical engineering, it has been used for the synchronization and monitoring of batch processes [9]. DTW has been effectively used to align biometric data, such as gait [10], signatures [11], fingerprints [12], and ECGs [13]. Rath and Manmatha have successfully applied DTW to the problem of indexing repositories of handwritten historical documents [14] and is also used in indexing video motion streams [15]. In robotics, Schmill et al. demonstrate a technique that utilizes DTW to cluster robots sensory outputs [17]. And in music, Zhu and Shasha have exploited DTW to query music databases with snippets of hummed phrases [18,19].

The superiority of DTW over Euclidean distance for these tasks has been demonstrated by many authors [8, 6, 13]

However, the greater accuracy of DTW comes at a cost. Depending on the length of the sequences, DTW is typically hundreds or thousands of times slower than Euclidean distance. But, extensive research has been performed on how to accelerate DTW computations, in particular for one-dimensional (or low-dimensional) real-valued sequences, often referred to as time series.
The main contributions of this paper are summarized as follows:

- We investigate the DTW technique in clustering time dependant sequences. Compared to the previous work [1], DTW technique provide better accuracy in formed clusters than Euclidian metric;
- We evaluate DTW performance in term of speed. Compared to Euclidian metric, DTW is many of times slower than Euclidean distance. In spite of that the global time taken by the clustering process with DTW technique is less many of times than the time taken without clustering;
- We also evaluate the generalization ability of DTW in clustering cross other databases.

The remainder of this paper is structured as follows. The two next sections briefly review the HMM on which we apply our proposed technique and the Dynamic time warping technique. Section 4 describes the proposed technique for reduced time learning algorithm. Obtained results are discussed in section 4. Section 5 discusses the possibility of an extended usability of the proposed technique on other database. Finally, section 6 concludes the paper.

## 2. HIDDEN MARKOV MODELS[24]

A Hidden Markov Model is a model for sequential data. It is a doubly embedded stochastic process with an underlying stochastic process that is not observable (hidden), but can only be observed through another set of stochastic process that produce the sequence of observations. The performance of a generative model like the HMM depends heavily on the availability of an adequate amount of representative training data to estimate its parameters, and in some cases its topology.

### 2.1 Elements of an HMM:

An HMM is characterized by the following:

1- N : the number of hidden states in the model. For many practical applications there is often some physical significance attached to the states or to the sets of states of the model. We denote the individual state as S= {S1, S2…Sn} and the state at time t as qt.

2- M : the number of distinct observation symbols. The observation symbols correspond to the physical output of the system being modeled. We denote the individual symbols as V={V1, V2…VM}

3- The state transition probability distribution A={aij}where: aij=P[qt+1=Sj|qt=Si]    1<=i, j<=N.

4- The observation symbol probability distribution in state j,  B={bj(k)}, where:
Bj(k)=P[Vk at t|qt=Sj]  1<=j<=N; 1<=k<=M.

5- The initial state distribution $\pi$ ={ $\pi$i} where $\pi$ =P[q1=Si]    1<=i<=N.
Given appropriate values of N, M, A, B and $\pi$ the HMM can be used as generator to give an observation sequence : O= O1O2….OT.
The compact notation: $\lambda$ =(A,B,$\pi$) is used to indicate the complete parameter set of the model.

## 2.2 The Three Basic Problems for HMMs:

Given the form of HMM there are three basic problems of interest that must be solved for the model to be useful in real world applications. These problems are the following:

1- Given the observation sequence $O=O_1O_2\ldots O_T$, and a model $\lambda =(A,B,\pi)$ how do we efficiently compute $P(O/\lambda)$, the probability of the observation sequence given the model?
2- Given the observation sequence $O=O_1O_2\ldots O_T$, and a model $\lambda =(A,B,\pi)$ how do we choose a corresponding state sequence $Q=q_1q_2\ldots q_T$ which is optimal in some meaningful sense .
3- How do we adjust the model parameters $\lambda=(A,B,\pi)$ to maximize $P(O/\lambda)$?

Problem three is the crucial one for most applications of the HMMs, since it allows us to optimally adapt model parameters to observed training data to create best models for real phenomena. The iterative procedure like Baum-Welch method or equivalently the EM (Expectation, maximization) [25] method is used to solve this problem. One associated problem to this solution is the time taken by the training process. In this paper, we aim to reduce training time by reducing the original training set via an unsupervised clustering. Any information loss during clustering is performed, we introduce a weight factor that is assigned to each training instance which represent each cluster, and modify all expectation formulas to include the weight factor in appropriate terms. The weight factor is the "frequency" of observing an instance with similar behavior (characteristics) in the training set. Consequently the obtained model parameters: A, B, $\pi$ are equal to those obtained with the classical HMM training algorithm but in less time.

## 3. DYNAMIC TIME WARPING (DTW)

DTW algorithms have been proposed around 1970 in the context of speech recognition, when it was mainly applied to isolated word recognition [25, 29].
DTW is a well-known technique to find an optimal alignment between two given (time-dependent) sequences:
$$X :=(x1, x2, \ldots , xN) \text{ of length } N \in \mathbb{N} \text{ and;}$$
$$Y := (y1, y2, \ldots , yM) \text{ of length } M \in \mathbb{N}$$
which may vary in time or speed. These sequences may be discrete signals (time-series) or, more generally, feature sequences sampled at equidistant points in time.
To align these two sequences using DTW, one needs a local cost measure, sometimes also referred to as local distance measure : $C(x,y)$.
Typically, $C(x, y)$ is small (low cost) if x and y are similar to each other, and otherwise $C(x, y)$ is large (high cost). Then, the cost matrix $C(N,M) \in R^{N\times M}$ defined by $C(n,m) := C(x_n, y_m)$ is constructed. To find the best match between the two sequences, we can find a path through the obtained matrix. A warping path, p, is a contiguous set of matrix elements that characterizes a mapping between X and Y by assigning the element $x_{nl}$ of X to the element $y_{ml}$ of Y .
The total cost $C_p(X, Y )$ of a warping path p between X and Y with respect to the local cost measure c is defined as:

$$c_p(X,Y) := \sum_{\ell=1}^{L} c(x_{n_\ell}, y_{m_\ell}).$$

Furthermore, an optimal warping path between X and Y is a warping path p* having minimal total cost among all possible warping paths. The DTW distance DTW(X, Y ) between X and Y is then defined as the total cost of p*:

$$\begin{aligned}\mathrm{DTW}(X,Y) &:= c_{p^*}(X,Y) \\ &= \min\{c_p(X,Y) \mid p \text{ is an } (N,M)\text{-warping path}\}\end{aligned}$$

## 4. TRAINING SET CLUSTERING USING DTW

In many real world classification problems there is significant data redundancy in training data, consequently, the time and memory complexity grows with sequence length, the number of training sequences, and with the number of HMM states N. For such problems one might be able to reduce the learning time by preprocessing the data.

Training HMM means optimizing the model parameters (A,B,π) to maximize the probability of the observation sequence P(O|λ). In order to obtain a good estimation of HMM; a big amount of computation and a large database is required.

HMM profiles are usually trained with a set of sequences that are known to belong to a single category. Consequently this set of sequences is often redundant in the sense that most of these sequences have the same values taken by the different features describing this category.

Many real world applications are sensitive to these phenomena as instance, in medicine, and for many patients reached with the same disease can have the same symptoms associated with this disease. In gender recognition or skin recognition, many subjects can have same characteristics (color, texture…). In facial expression recognition, many subjects display the same description of any facial expression (e.g Joy : the distance between mouth corners increases, the distance between lips relaxes or increases and the distance between lids decreases or relaxes) [26].

In image analysis, in many times, the extracted wavelets or FFT coefficients from some image parts are the same when the images are belonging to the same category, etc…

Our aim is to reduce the size of the training set, so, the time of the training process without reducing generalization (classification) accuracy. In this context, our approach involves a preprocessing step to reduce the number of training samples, so we do not require that all data to be resident in main memory (minimize storage and time training requirements). To this end, we apply an unsupervised clustering on the original training data set to extract representative clusters. To preserve training data from any information loss, we associate a weight factor to each cluster. This weight factor is the frequency of observing instances with similar behavior (characteristics). This approach is suitable for many real world applications where there is significant data redundancy. After that, we modify all expectation formulas computed with the classical HMM learning method (EM) to include the weight factor in appropriate terms.

The proposed approach reduces significantly the run-time of HMM training and the storage requirement, while providing accurate discovery and discrimination of different category behaviors when applying the unsupervised clustering algorithm.

## 4.1 Clustering Process

Given an original training set, we apply an unsupervised clustering on all training instances that belong to a particular class. That is, unsupervised clustering is performed independently on each class. There are numerous clustering techniques including the K-means, fuzzy C-means, hierarchical clustering, and self-organizing maps. The common problem with most of these methods is the choice of the number of clusters or the number of neighbors. In our approach we propose a simpler clustering algorithm where, it is important to note that the number of clusters is not known a priori.

*Clustering algorithm:*
Begin
N: is the number of categories;
ni: is the number of instances (sequence observation) associated to category i;   i=1..N
$C_k(i)$: cluster number k of category i;           1<=k<=ni;   j=sequence length
&
$C_k(i,j+1)$= frequency of sequence observation
(instance with same features values)

For all categories(N)
   Put the first instance from the training set of the category i in table C of clusters;
   Initialize the frequency field by 1;
   For all instances belonging to each category(ni)
      Compare the new instance(inst_nouv) with all instances (inst_in) in the cluster table;
      If  DTW(Inst_nouv,inst_in)=0
         Update frequency field(+1);
      Else
         Add a new cluster with the new instance;
         Initialize the frequency field with 1;
      End
   End
End
End.

In our previous work [1], the Euclidian distance was used to compare two sequences. In this paper we investigate the DTW technique to compare sequences. Indeed, with Euclidean distance, the two compared sequences must have the same length, only corresponding Time Points are compared, but: In many real world applications, time dependant sequences are out-of-synchronization and produced sequences are not always in the same length. As instance in our experiments when studying facial expressions, one can display an expression (Joy) slowly and another one displays the same expression (Joy) rapidly. Hence, the first recorded video produces more frames than the second one, so the two sequences have not the same length, also, facial features deformations in the first video (which are the same in the two videos) are not synchronized with those of the second one. In spite of that the two sequences can easily be aligned then clustered in the same cluster because both represent the same expression (Joy).
In another hand, generally, when modelling a situation with HMM, produced sequences count multiple redundant consecutive emitted observations, because of the structure of HMM which can stay in the same state or transit to another one and emit a set of observations with given probabilities.

International Journal of Data Mining & Knowledge Management Process (IJDKP) Vol.3, No.2, March 2013

Example:

Suppose we have two HMM with Two states. The possible observations which can be emitted by each state are: O={1,2,3,4,5,6,7,8}. Some possible produced sequences from the two HMMs can be: {1234567, 1222234, 1123334, 1222344...} as we can see, the first sequence represents one occurrence of each observation; the next ones represent multiple occurrences of different observations.

When applying Euclidean metric to cluster these sequences, we get four clusters, but when using DTW technique we get only two clusters, the distance between the three last sequences is equal to "0". Consequently, the first sequence can be assigned to the first HMM and the three other sequences can be assigned to the second HMM. The maximum likelihood confirms the decision. Thus, DTW technique is most accurate than the Euclidean distance since it produces meaningful clusters.

Once clustering is completed, we use each instance in the cluster table as a new training instance. Its weight factor can be found in the associated frequency field.

In many studies, clusters with small weight are considered as noisy instances so they are removed. It is not the case in all studies, in our experiments, each cluster describe a specific behavior associated to the considered class (e.g Anger or disgust are two universal facial expressions which can be displayed in different ways, each cluster of each class (facial expression) can be a specific description of each facial expression so it is not a noisy instance), in this case only a human expert can remove these clusters to improve classification accuracy.

## 4.2 The Modified EM Algorithm for HMM Training

Generally, the training of Hidden Markov Models is done by the EM Algorithm to create the best fitting HMM from a given set of sequences F.

The EM algorithm for HMMs
- Initialization : set initial parameters $\lambda^0$ to some value;
- For t = 1..num iterations
    - Use the forward_backward algorithm to compute all expected counts :
      Number of transitions from state $S_i$ to $S_j$ using the precedent set of parameters $\lambda^{t-1}$
      Number of emissions of symbol $V_k$ from state $S_j$ using the precedent set of parameters $\lambda^{t-1}$
      Update the parameters based on the expected counts using formulas (2) and (3).

End.

From the initial HMM M and the initial parameter set $\lambda$ we can generate a new optimized parameter set $\lambda'$. Using the concept of counting event occurrences, the following formulas are given for re_estimation of the new HMM parameters:
- $\pi_i$'= The expected frequency in state $S_i$ at time t=1 **(1)** ;
- $a_{ij}$'= the expected number of state $S_i$ to state $S_j$ transitions**/** the expected number of transitions from state $S_i$ **(2);**
- $b_j(k)$'= The expected number of times the observation $V_k$ is generated from state $S_j$ **(3);**

These computations are performed for all training instances, in our case, these computations are performed for the new set of training instances, it consists on the clustered training set instances so they are iterated for m=1..$k_i$ ($k_i$ is the number of clusters of class i), multiplying each formula by $C_k(i,j+1)$ (number of similar instances);

As a result the transition probability from state $S_i$ to state $S_j$ and the emission probability of the observation $V_k$ from state $S_j$ increase with the number of instances (redundant and non redundant). More than that, time running in the computation of all these expectations and for all iterations is significantly reduced.

## 5. EXPERIMENTAL RESULTS

The experiments were conducted on both (i) extracted information from (MMI database[27]) videos and (ii) synthetically generated instances. Our aim is to analyze the capability of the proposed method in reducing training set instances without information loss and reducing training time. Our code source is written in Matlab, and, our experiments are done on: Intel Dual-Core 1.47GHz machine with 2G memory.

### 5.1. MMI database

The MMI database is a resource for Action unit (AU) and basic emotion recognition from face video. It consists of V parts; each part is recorded in different conditions. The second set of data recorded was posed displays of the six basic facial expressions: Joy, Sadness, Anger, Disgust, Fear and Surprise. In total, 238 videos of 28 subjects were recorded. All expressions were recorded twice. In our experiments [28], we aim to discover novel clusters with different behavior relative to each class (generate for each facial expression a set of clusters which allow describing all possible behaviors associated with each studied class). To this end, facial characteristic points were detected on the first frame of each video and tracked, then, characteristic distances were computed leading to several time series. Obtained time series were analyzed to produce a smaller set of time series which represent occurrence order of facial features deformations of each subject with each facial expression (This is what we call data preparation).

### 5.2. Synthetic Time Series

To generate synthetic data, the Markov Model with its transition and emission probabilities matrixes constructed with the MMI database was used. Accordingly, for a specific alphabet size M=10 observations, several training sequences were produced.

With both data, we evaluated our experiments on an HMM with three states(3), sequence length equal to five (5) from where we have extracted the most interesting subsequences which length is three (3), the number of possible observations is ten (10) and the number of iterations is fifty (50). The estimated time training of the EM and EM based cluster training algorithms are presented on table 1 according to each experiment:

Table 1. Time training of EM and EM based cluster training algorithms.

| Number of sequence observations | Number of clusters | Time of clustering /s | | Time of EM training/s | Time of EM based cluster training/s | DIFF in number of times |
|---|---|---|---|---|---|---|
| | | Euclidean distance | DTW technique | | | |
| **100** | 10 | 0 | 0.0780 | 0.5304 | 0.0936 | **5.6** |
| **1.000** | 19 | 0.0312 | 1.0764 | 5.234 | 0.162 | **32.3** |
| **10.000** | 24 | 0.2652 | 7.9717 | 44.3979 | 0.1945 | **228.26** |
| **100.000** | 25 | 4.6020 | 135.4245=2mn25s | 497.4092=8mn29s | 0.2106 | **2361.87** |

Results on this table are computed as the mean of ten (10) executions of the two algorithms. In term of processing speed, the EM training takes 0.5304s to learn 100 training sequences. With the EM based cluster training, it takes 0.0936s, it means that it is 5.6 times faster than EM training. When the number of training sequences is equal to 1.000, the EM based cluster training is 32.3 times faster than the EM training. When the number of training sequences is equal to 10.000, the EM based cluster training is 228.26 times faster than the EM training. Finally, when the number of training sequences is equal to 100.000, the EM based cluster training is 2361.87 times faster than the EM training.

However, we can observe that time clustering with Euclidean distance is less than time clustering with DTW technique in all cases. Anyway, we can explore one of the proposed methods presented in the literature to speed up DTW computation. The complexity of this technique is $O(N.M)$, it can be reduced to $O(N+M)$ like in [30].

With conducted experiments on these databases, we can observe that the number of constructed clusters when using Euclidean metric is the same when using DTW technique this can be explained by the fact that the new training set sequences (representative/index clusters) are really different (no multiple occurrences of some observations in the same sequence). To evaluate the ability of DTW in the construction of more accurate fewer clusters, we use another database (see section 5).

Although clustering based DTW technique consumes more time than clustering based Euclidean metric, the overall time learning (clustering + training) based clustering with Euclidean/DTW is less than time learning without clustering.

The experiments show that the difference in time training between the two algorithms increases dramatically with the number of training sequences in favor of the proposed modified EM training algorithm (it reduces). These results suggest that the proposed method can be used especially with very large datasets.

Obtained results show that the proposed process reduces not only time computation requirements, but, it also reduces significantly storage requirements. Because HMM profiles are usually trained with a set of sequences that are known to belong to a single category, the clustering process produce a small number of clusters. These clusters form indexes of constructed clusters and will represent the new set of training instances.

Experimental results show that both EM and EM based cluster training algorithms achieve similar: Initial (π), Transition (A) and Emission (B) probabilities matrixes. Consequently the classification accuracy is maintained.

## 6. GENERALIZATION TO OTHER DATABASE

To evaluate the ability of DTW in more accurate clustering, we conduct our experiments on another database. The new database count 30.000 sequences obtained from MMI facial expressions database, the length of each sequence is about nine (9) observation instead of three and many sequences have multiple occurrences of the same observation. The evaluation of the proposed approach on this database is summarized on table 2:

Table 2. Time clustering, time training of EM and EM based cluster training algorithms with Euclidean and DTW distances.

| Number of Sequences | Number of real clusters | Euclidean distance | | | DTW technique | | | Time Training without clustering |
|---|---|---|---|---|---|---|---|---|
| | | Number of constructed clusters | Time Clustering | Time Training based clustering | Number of constructed clusters | Time Clustering | Time Training based clustering | |
| 30.000 | **70** | 70 | 2.26 | 0.41 | 25 | 79.7477 | 0.18 | 163.45 |

As we have explained before, we can note that the number of constructed clusters when using DTW is less than the one obtained when using Euclidean metric, it match very well the real number of clusters. This can be explained by the fact that many sequences with Euclidean metric are considered as different in spite that they are similar but not synchronized. This means that DTW are more accurate in the field of clustering than Euclidean metric.

Also, we can observe that time training based clustering with DTW is less than the one with Euclidean metric, this can be explained by the number of constructed clusters which is greater when using Euclidean metric.

Finally we can say that in general, the proposed idea can be adapted to all ML methods which involve a number of iterations equal to the number of instances in their processing. The proposed process reduces the number of iterations and consequently reduces the time training of these ML methods.

## 7. CONCLUSION

In this paper, a more accurate simple process to speed up ML methods is studied. An unsupervised clustering algorithm is combined with EM algorithm to develop an efficient HMM training to tackle the problem of HMM time training of large datasets. The idea of the proposed process consists of two steps. In the first step, training instances with similar inputs are clustered in one cluster and each cluster is weighted with the number of similar instances. DTW technique is used to accurately cluster training time dependant sequences. In the second step, all formulas associated with the number of training instances are modified to include the weight factor in appropriate terms. Through empirical experiments, we demonstrated that the EM based cluster training algorithm reduces significantly the HMM time training and storage requirements. Furthermore, the proposed approach is not limited to HMMs it can be employed for a large variety of ML methods. In a future work, we plan to explore new techniques to cluster more accurately time dependant sequences, to reduce more and more time training and improve more and more classification accuracy and test these methods on different ML methods.